# A Survey on Soft Subspace Clustering

Zhaohong Deng[1,2], Kup-Sze Choi[3], Yizhang Jiang[1], Jun Wang[1], Shitong Wang[1]

[1] *School of Digital Media, Jiangnan University, Wuxi, Jiangsu, P.R. China*
[2] *Department of Biomedical Engineering, University of California, Davis, U.S.A.*
[3] *The Centre for Smart Health, School of Nursing, The Hong Kong Polytechnic University, Hong Kong*

**Abstract:** Subspace clustering (SC) is a promising technology involving clusters that are identified based on their association with subspaces in high-dimensional spaces. SC can be classified into hard subspace clustering (HSC) and soft subspace clustering (SSC). While HSC algorithms have been studied extensively and are well accepted by the scientific community, SSC algorithms are relatively new. However, as they are said to be more adaptable than their HSC counterparts, SSC algorithms have been attracting more attention in recent years. A comprehensive survey of existing SSC algorithms and recent developments in the field are presented in this paper. SSC algorithms have been systematically classified into three main categories: conventional SSC (CSSC), independent SSC (ISSC), and extended SSC (XSSC). The characteristics of these algorithms are highlighted and potential future developments in the area of SSC are discussed. Through a comprehensive review of SSC, this paper aims to provide readers with a clear profile of existing SSC methods and to foster the development of more effective clustering technologies and significant research in this area.

* Corresponding author: zhdeng@ucdavis.edu, dzh666828@aliyun.com

# 1 Introduction

Despite extensive studies of clustering techniques over the past decades, in various areas of application such as statistics, machine learning, and databases [40,69], conventional techniques fall short when clustering is performed in high-dimensional spaces [44,52,59]. A key challenge of most clustering algorithms is that, in many real-world problems, data points in different clusters are often correlated with some subsets of features, i.e., clusters may exist in different subspaces or in a certain subspace of all features. Therefore, for any given pair of neighboring data points within the same cluster, it is possible that the points are indeed far apart from each other in a few dimensions of high-dimensional space.

In recent years a plethora of subspace clustering (SC) techniques have been developed to overcome this challenge. The goal of SC is to locate clusters in different subspaces or in a certain subspace of the original data space. The two main classes of SC algorithms are Hard Subspace Clustering (HSC) and Soft Subspace Clustering (SSC). Research into SC begins with an in-depth study of HSC methods for clustering high-dimensional data. With HSC algorithms, an attempt is made to identify the exact subspaces for different clusters, a process that can be further divided into *bottom-up* and *top-down* subspace search methods [52]. Examples of the former are CLIQUE [3], ENCLUS [17], and MAFIA [35]; and that of the latter are ORCLUS [1], FINDIT [65], DOC [57], $\delta$-Clusters [70], and PROCLUS [2]. Other common HSC algorithms include HARP [71] and LDR [10]. A detailed review of HSC algorithms can be found in [44,45,52,59].

While the goal of HSC is to identify exact subspaces, SSC algorithms perform clustering in high-dimensional spaces by assigning a weight to each dimension to measure the contribution of individual dimensions to the formation of a particular cluster. SSC can be considered an extension of conventional *feature weighting clustering* [9,18,21,22,39,47,50,51,61,64]. In this paper, SSC algorithms are hierarchically classified into three main categories: (1) *conventional SSC* (CSSC), (2) *independent SSC* (ISSC), and *extended SSC* (XSSC). Here, CSSC refers to conventional feature weighting clustering algorithms, i.e., where all clusters share the same subspace and a common weight vector. By contrast, the weight vectors in ISSC are different for different clusters. In other words, each cluster has an *independent* subspace. Thus, ISSC can also be referred to as *multiple features weighting clustering*. XSSC represents a category of algorithms that were developed by extending CSSC or ISSC algorithms by introducing new mechanisms to enhance clustering performance or for some specific purposes. The definitions of these three types of SSC are given in Table 1. A detailed review of the characteristics of the algorithms will be carried out.

Compared with traditional non-subspace clustering techniques, SSC has demonstrated promising performance in data clustering, especially for high-dimensional data. Through a comprehensive review of SSC, the aim of this paper is to provide readers with a clear profile of existing SSC methods and to foster the development of more effective technologies and significant research in this area.

The rest of this paper is organized as follows. The classification of existing SSC algorithms is given in Section 2. The CSSC, ISSC, and XSSC algorithms are comprehensively reviewed in Sections 3, 4, and 5, respectively. The algorithms are compared and analyzed theoretically in Section 6, and experimentally in Section 7. Conclusions are given in Section 8. For clarity, the notations used in this paper are defined in Table 2.

**Table 1 Definitions of three categories of SSC algorithms**

| SCC algorithms | Descriptions |
|---|---|
| Conventional SSC (CSSC) algorithms | Classic feature weighting clustering algorithms, with all of the clusters sharing the same subspace and a common weight. |
| Independent SSC (ISSC) algorithms | Multiple feature weighting clustering algorithms, with all of the clusters having their own weight vectors, i.e., each cluster has an independent subspace, and the weight vectors are controllable by different mechanisms. |
| Extended SSC (XSSC) algorithms | Algorithms extending the CSSC or ISSC algorithms with new clustering mechanisms for performance enhancement and special purposes. |

Table 2 Notations commonly used by the algorithms discussed in this paper

| Notations | Description |
|---|---|
| $\mathbf{U}=[u_{ij}]_{C\times N}$ | Hard/fuzzy partition matrices |
| $\mathbf{V}=[\mathbf{v}_1,\cdots,\mathbf{v}_C]^T$, $\mathbf{v}_i=[v_{i1},\cdots,v_{iD}]^T$ | Clustering centers matrix, where $\mathbf{v}_i$ is the center of the $i$ th cluster. |
| $\mathbf{W}=[w_{ij}]_{C\times D}=[\mathbf{w}_1,\cdots,\mathbf{w}_C]^T$, $\mathbf{w}_i=[w_{i1},\cdots,w_{iD}]^T$ | Weighting matrix $\mathbf{w}_i$, where the weight vector is associated with the $i$ th cluster. |
| $m$ | Fuzzy index of fuzzy memberships |
| $\tau$ | Fuzzy index of fuzzy weighting |
| $C^*$ | Number of clusters |
| $T$ | Total number of iterations for iteration-based algorithms |
| $N$ | Number of data/samples |
| $D$ | Number of features |

* For simplicity and consistency, we have used the same notation C to denote the number of clusters, although K is commonly used to denote the number of clusters for K-means type algorithms.

## 2 Classification of SSC

As discussed previously, SSC algorithms can be broadly classified into three main categories: CSSC, ISSC, and XSSC. Each of these categories can be further divided into subcategories based on the clustering mechanisms that are adopted, as shown in Table 3. In CSSC, clustering is performed by first identifying the subspace using some strategies, and then carrying out clustering in the subspace that was obtained, in order to partition the data. This is referred to as *separated feature weighting*, where data partitioning involves two separate processes – subspace identification and clustering in subspace. Clustering can also be conducted by performing the two processes simultaneously, an approach known as *coupled feature weighting*. In ISSC, algorithms are developed based on the K-means model, fuzzy C-means (FCM) model, and probability mixture model, in a process where fuzzy weighting, entropy weighting, or other weighting mechanisms are adopted to implement feature weighting. Finally, XSSC algorithms can be subdivided into eight subcategories, depending on the strategies used to enhance the CSSC and ISSC algorithms. These subcategories are between-class separation, evolutionary learning, the adoption of new metrics, ensemble learning, multi-view learning, imbalanced clusters, subspace extraction in the transformed feature space, and other approaches such as the reliability mechanism and those used for clustering categorical datasets. The classification of SSC algorithms is presented in Table 3. These algorithms will be discussed in the following sections.

**Table 3 Classification of SSC algorithms**

| CSSC | | ISSC | | | |
|---|---|---|---|---|---|
| Separated feature weighting | Coupled feature weighting | K-means model | FCM model | Probability mixture model | Others |
| [22,50,51] | [9,18,21,39,47,61] | [11,25,26,32,33,41,42] | [29,43] | [13,14,54] | [28] |
| XSSC | | | | | | | |
|---|---|---|---|---|---|---|---|
| Between-class separation | Evolutionary learning | Adaptive metric | Ensemble learning | Multi-view learning | Imbalanced data learning | Transformed feature space | Others |
| [20,36,46] | [34,49,72] | [19,58,63] | [24,37] | [12,16,30] | [53] | [23,60] | [4,6,8,15,48,67] |

## 3 CSSC

CSSC algorithms can be classified based on the strategies of the feature weighting that are adopted. Thus, CSSC algorithms are divided into two categories: those that adopt separated feature weighting and those that employ coupled feature weighting. In the former, the weights are determined before clustering is performed; while in the latter the weights are learned during the clustering process. In Table 4, the representative CSSC algorithms are listed with brief descriptions. More details of these algorithms are presented below.

**Table 4 Descriptions of Nine CSSC Algorithms**

| Algorithm | Objective function of clustering | Method of feature weighting | Strategy of feature weighting* |
|---|---|---|---|
| *(1) C-K-means* | Weighting K-means | *Separated Feature Weighting* | (1) A set of feasible weight groups are first defined as candidates for optimal weights before data clustering is carried out; (2) For each candidate weight group, the k-means algorithm based on this weight group is used to generate a data partition; (3) The Fisher ratio calculated from the partition that is obtained based on different weight groups is adopted as the evaluation index to determine the optimal weight group. |
| *(2) OVW-UAT* | Weighting K-means | | (1) Two objective functions with the weights as the variables are proposed to determine the weights; (2) K-means clustering is then implemented based on the feature weights that were obtained. |
| *(3) WFCM* | Weighting FCM | | (1) A specific objective function with the weights as the variables is given to obtain the optimal weight; (2) The objective function is then solved using the gradient descent learning technique; (3) The FCM is implemented based on the feature weights that were obtained. |
| *(4) SYNCLUS* | Weighting K-means | *Coupled feature weighting* | (1) Feature weighting is implemented by solving a specific objective function; (2) K-means is implemented based on the feature weights that were obtained; (3) The above two steps are alternately implemented. |
| *(5) FWSA* | Weighting K-means | | (1) A weighting k-means objective function is defined to optimize partitioning; (2) An objective function for optimizing the weights is also defined; (3) The partition and the weights are updated by solving the above two objectives in an iterative manner. |
| *(6) W-k-means* | Weighting K-means | | (1) A weighting k-means objective function is defined; (2) Based on the above objective function, rules for updating the partition and the weights are obtained; (3) The partition and the weights are updated in an iterative manner using the updating rules. |
| *(7) FWFKM* | Weighting FKM | | (1) The fuzzy k-means algorithm is used to obtain the partition based on the weights that were given; (2) Based on the partition the supervised ReliefF algorithm is used to obtain the feature weights; (3) The above steps are implemented alternately. |
| *(8) MWLA* | Weighting Gaussian mixture model | | (1) The Maximum Weighted Likelihood (MWL) learning framework in the context of the Gaussian mixture model is proposed; (2) The cluster's structure and the relevant features are learned automatically and simultaneously in an iterative manner. |
| *(9) MA-DDC-FW* | Probabilistic model | | (1) A probabilistic model is used to define the objective for clustering; (2) Based on the above model, the relevance weights and the partition are learned in an iterative manner. |

* The term "strategy of feature weighting" refers to the mechanism that is adopted to realize the assignment of weight values for each feature.

### 3.1 Separated Feature Weighting Algorithms

#### (1) C-K-means

The Convex K-means (C-K-means) is a method proposed specifically for variable weighting in k-means clustering [51]. The aim of this method is to optimize the weights of the variables in order to achieve the best clustering result by minimizing the generalized Fisher ratio $Q$ – the ratio of the average within-cluster distortion to the average between-cluster distortion. To find the minimum $Q$ value, a set of feasible weight groups is first defined. For each weight group, the k-means algorithm is used to generate a data partition and the $Q$ value is calculated for the partition. The most desirable cluster is then given by the partition with the minimum $Q$ value. The shortcoming of this method is that the optimal weights are not guaranteed in the predefined set of weights. Also, it is not practical to obtain a predefined set of weights for high-dimensional data.

#### (2) OVW-UAT

The Optimal Variable Weighting for Ultrametric and Additive Tree (OVW-UAT) clustering is a feature weighting strategy that was developed to allow hierarchical clustering methods to be used to solve variable weighting problems [22]. In this approach, two objective functions are used to determine the weights for trees in ultrametric and additive forms, respectively. Once the optimal variable weights are obtained, the resulting inter-object dissimilarities can be applied to any of the existing ultrametric or additive tree fitting procedures. Since hierarchical clustering methods are computationally complex, the OVW-UAT approach cannot handle large datasets efficiently. Makarenkov and Legendre extended the OVW-UAT approach to optimal variable weighting for k-means clustering [50]. The simulation results showed that the method was effective for identifying important variables. Compared with the abovementioned C-K-means, the advantage of OVW-UAT is that the weighting can be optimized by optimizing the corresponding objective function, while with C-K-means an optimal solution can only be found within the given weight groups. However, this algorithm is still not scalable to large datasets.

#### (3) WFCM

The Weighted FCM (WFCM) algorithm is another clustering method that is grouped under the category of separated feature weighting [64]. In the algorithm, clustering is performed by employing the weighted Euclidean distance as a metric, incorporating feature weights into the commonly used Euclidean distance. The algorithm begins by estimating the weight vector using the objective function below and the gradient descent learning technique,

$$\min\ E(\mathbf{w}) = \frac{2}{N(N-1)} \sum_{i=1}^{N} \sum_{j=1, j \neq i}^{N} \frac{1}{2}\left(\rho_{ij}^{(\mathbf{w})}(1-\rho_{ij}) + \rho_{ij}(1-\rho_{ij}^{(\mathbf{w})})\right),$$

$$\rho_{ij}^{(\mathbf{w})} = \frac{1}{1 + \beta \cdot d_{ij}^{(\mathbf{w})}}\ ,\quad d_{ij}^{(\mathbf{w})} = \sqrt{\sum_{k=1}^{D} w_k^2 (x_{ik} - x_{jk})^2}\ ,$$

where $E(\mathbf{w})$ is a function of weighting variable $\mathbf{w}$ for obtaining the optimized weights; $N$ is the number of samples; and $d_{ij}^{(\mathbf{w})}$ and $\rho_{ij}^{(\mathbf{w})}$ denote $d_{ij}$ and $\rho_{ij}$ in the original space, respectively. Once the weights are determined, WFCM can be implemented by replacing the common Euclidean distance in FCM with the weighted Euclidean distance. As with the OVW-UAT algorithm, the feature weights in WFCM can be optimized by optimizing the corresponding objective functions. Nevertheless, three algorithms, i.e., C-k-means, OVW-UAT, and WFCM, have a common limitation − the feature weights must be determined before the clustering procedure is carried out. Thus, the feature weights cannot be further optimized in the subsequent clustering procedure.

### 3.2 Coupled Feature Weighting Algorithms

Coupled feature weighting CSSC algorithms are reviewed in this subsection. They are different from the algorithms discussed in Subsection 3.1 in that the feature weights can be updated adaptively in the clustering process.

*(1) SYNCLUS, FWSA, and FWFKM*

The Synthesized Clustering (SYNCLUS) algorithm is developed to deal with variable weighting in k-means clustering [21]. The algorithm consists of two stages. Starting with an initial set of weights, SYNCLUS first employs k-means clustering to partition data into $k$ clusters. This is followed by the estimation of a set of new weights for different features, performed by optimizing a weighted mean-square stress-like cost function. The two stages proceed iteratively until convergence to an optimal set of weights is achieved. SYNCLUS can effectively optimize the weights of features and the partitioning of data simultaneously by optimizing two objective functions alternately. However, this algorithm is computationally intensive and very time-consuming [68], making it unsuitable for the handling of large datasets.

Tsai and Chiu proposed the Feature Weight Self-Adjustment Algorithm (FWSA), which is based on the k-means clustering model [61]. The algorithm adopts the objective function below:

$$\min\ J_{FWSA}(\mathbf{U},\mathbf{V}) = \sum_{i=1}^{C}\sum_{j=1}^{N} u_{ij} \sum_{k=1}^{D} w_k (x_{jk} - v_{ik})^2$$

$$\text{s.t.}\ \ u_{ij} \in \{0,1\},\ \sum_{i=1}^{C} u_{ij} = 1,\ 0 \le w_k \le 1,\ \text{and}\ \sum_{k=1}^{D} w_k = 1.$$

Furthermore, the following sub-optimization function is used to adjust the weights

$$\max\ E(\mathbf{w}) = \frac{\sum_{i=1}^{C} \|C_i\| \sum_{k=1}^{D} w_k (v_{ok} - v_{ik})^2}{J_{FWSA}(\mathbf{w})}\ ,$$

where $\|C_i\|$ denotes the number of the data objects in the $i$th cluster obtained in the current iteration and $\mathbf{v}_o = [v_{o1},\cdots,v_{oD}]^T$ is the global center of all data objects in the dataset. The final clustering results and weights are then obtained iteratively. Like SYNCLUS, two objective functions are needed to optimize the partitioning of data and the weights of features, respectively. This makes it difficult to rigorously analyze the convergence of the algorithm.

The Feature Weighted Fuzzy K-means (FWFKM) algorithm performs clustering through an iterative procedure based on the fuzzy k-means algorithm and the supervised ReliefF algorithm [47]. Suppose that $D$ is the number of features. The FWFKM algorithm begins by setting the weights as $1/D$ and implementing the fuzzy k-means algorithm with the weighted distances to obtain the initial clustering result and label the data. With the labeled data, the supervised ReliefF algorithm is then used to assign new weights for every feature. This procedure is conducted iteratively, with the weights updated repeatedly until the final clustering result is achieved. Since ReliefF is a classic technique of supervised feature weighting, FWFKM can effectively realize supervised learning in the clustering procedure with the labeled data obtained in the last partitioning of data by fuzzy k-means. However, two objective functions are still needed for FWFKM to optimize the partitioning of data and the weights of features, respectively.

*(2) W-k-means*

For the three coupled feature weighting algorithms discussed above, i.e., SYNCLUS, FWSA, and FWFKM, we see that the processes of partitioning data and learning feature weights are implemented in an alternate manner by optimizing the respective objective functions. This means that the convergence of the learning may be not guaranteed, since the objective functions are constantly changing throughout the whole learning process. In order to overcome this weakness, efforts have been made to develop algorithms that optimize the partitioning of data and the feature weights using a common objective function, such as the classic W-*k*-means algorithm, which is discussed below [39].

Huang et al. proposed W-*k*-means, a k-means type of automated variable weighting clustering algorithm, using the following objective function [39]:

$$\min J_{W-k-means}(\mathbf{U},\mathbf{V},\mathbf{w}) = \sum_{i=1}^{C}\sum_{j=1}^{N} u_{ij} \sum_{k=1}^{D} w_k^{\tau}(x_{jk}-v_{ik})^2$$

s.t. $u_{ij} \in \{0,1\}$, $\sum_{i=1}^{C} u_{ij} = 1$, $0 \leq w_k \leq 1$, $\sum_{k=1}^{D} w_k = 1$.

With the current partition in the iterative k-means clustering process, the W-*k*-means algorithm calculates a new weight for each variable, i.e., feature, based on the variance of the within-cluster distances. The new weights are used to decide the cluster memberships of the objects in the next iteration. The optimal weights are found when the algorithm converges, and these weights can then be used to identify important features for clustering. The features, which may be noise to the clustering process, can be removed in a future analysis. The convergence of W-*k*-means can be analyzed rigorously. This algorithm is receiving increasing attention, and many modified versions have been proposed, such as the fuzzy subspace clustering algorithm [25, 41], to be discussed in Subsection 4.1.1.

*(3) MWLA and MA-DDC-FW*

In the abovementioned CSSC algorithms, the classic K-means and FCM frameworks were adopted to develop the corresponding objective functions. Although these strategies have demonstrated good effectiveness, they do not adequately take into consideration the effect of data distribution. To address this weakness, probability models, e.g., the Gaussian mixture model, have been introduced to improve the corresponding CSSC algorithms.

Cheung and Zeng proposed the Maximum Weighted Likelihood (MWL) learning framework with the Gaussian mixture model to automatically and simultaneously identify the clustering structure and the related features [18]. The MWL-based algorithm (MWLA) is performed by introducing two sets of weight functions – one to reward each component in the mixture for its significance, and the other to discriminate between each feature of the clustering structure in terms of relevance. Thus, the MWLA can effectively consider the distribution information in the clustering procedure. However, since two functions are adopted for the components of the mixture and the features respectively, this algorithm is more complicated than the algorithms based on K-means and the FCM framework.

To address the problem of the unsupervised selection or weighting of discrete features, the Model-based Approach for Discrete Data Clustering and Feature Weighting (MA-DDC-FW) [9] is proposed, involving the use of a probabilistic approach to assign relevance weights to the discrete features. In the algorithm, the features are regarded as random variables modeled by finite discrete mixtures. Bayesian and information-theoretic approaches through stochastic complexity are both employed for the learning of the model. The feasibility and merits of MA-DDC-FW are well demonstrated in difficult problems involving clustering and the recognition of visual concepts in image data. The algorithm has also achieved success in text clustering.

### 3.3 Simulation on the Toy dataset

In this section, as one of the classic CSSC algorithms, W-$k$-means is adopted to illustrate the performance of the CSSC algorithm on a toy dataset. A toy dataset, denoted as Toy-D, with predetermined cluster structures is generated. Toy-D contains three clusters located at different subspaces, as shown in Fig. 1. The corresponding parameters used to generate the data are listed in Table 5. Each sub-figure in Fig. 1 corresponds to a cluster where the high-dimensional data are taken as the sequences and plotted in the corresponding sub-figures. In the sub-figures, the sequence number of the features and the value of the features are taken as the x-coordinate and y-coordinate, respectively. From these sub-figures, we can see the corresponding subspace of each cluster, the features of which are more important to the associated cluster than the other features. For example, from the three sub-figures in Fig. 1 and the predetermined parameters in Table 5, we can see that the feature values of most of the features of the three clusters are uniformly distributed in the interval [0, 100], but that the three clusters have distinctive feature subsets, with the sequence number of the features in the intervals of [1, 30], [20, 45], and [35, 55], respectively.

The performance of W-*k*-means on Toy-D is reported with Fig. 2 and Table 6. From Fig. 2, we can see that W-*k*-means can effectively assign larger values of weights to the important features. However, we also see that since all of the clusters share a common weight vector, the distribution of the weights that is obtained cannot effectively describe the importance of the features in each cluster. In Table 6, the clustering performance of W-*k*-means on Toy-D is reported with the means and standard deviations of the RI and NMI indices obtained by W-*k*-means with the ten runs, where RI and NMI are two commonly used indices to evaluate the clustering performance. Details of these two indices can be seen in Section 7. Both indices take a value within the [0, 1] interval. The higher the value, the better the clustering performance.

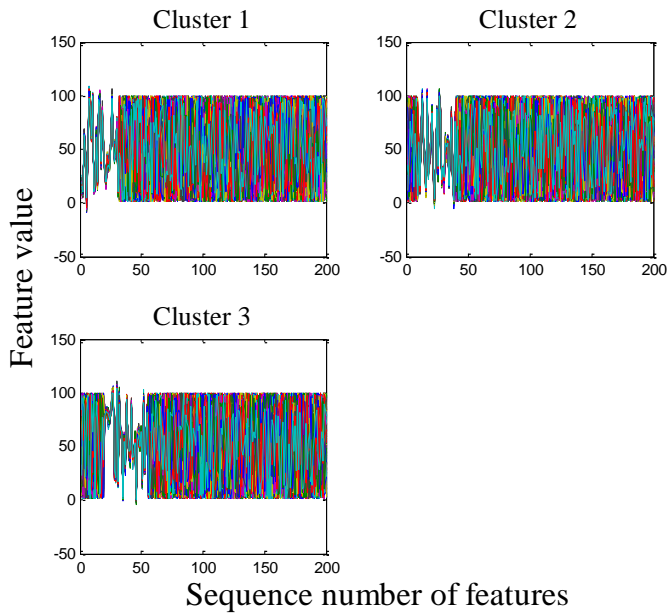

**Fig. 1 The dataset Toy-D.**

**Table 5 Parameters used to generate the Toy-D dataset.**

| | Synthetic dataset Toy-D | |
|---|---|---|
| Important attribute set | Cluster-1 | [1:30] |
| | Cluster-2 | [20:45] |
| | Cluster-3 | [35:55] |
| Size | Cluster-1 | 200 |
| | Cluster-2 | 200 |
| | Cluster-3 | 200 |
| Dimension | 200 | |

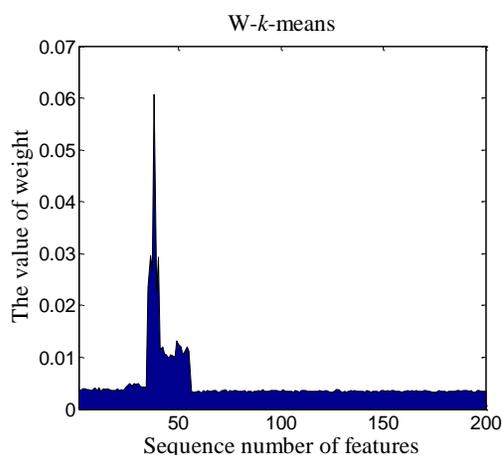

**Fig. 2 Distribution of weights obtained by the W-*k*-means algorithm on Toy-D**

**Table 6 Clustering indices obtained by W-*k*-means on Toy-D**

| Index | | W-*k*-means |
|---|---|---|
| *RI* | *mean* | 0.9708 |
| | *std* | 0.0117 |
| | * | 0.9842 |
| *NMI* | *mean* | 0.8727 |
| | *std* | 0.0327 |
| | * | 0.9080 |

\* Denotes the values of RI and NMI achieved by each algorithm when the lowest value of the loss function is obtained within the 10 runs.

## 4 ISSC

ISSC algorithms are distinct from CSSC in that each cluster has an independent subspace associated with a weight vector. The algorithms can be implemented based on the FCM/K-means model, the mixture model, and other models, and with the application of different approaches to weighting, such as fuzzy weighting, entropy weighting, and other mechanisms. For ISSC algorithms based on the FCM/K-means model and the fuzzy weighting mechanism, the parameters $w_{ik}^\tau$ are used for fuzzy weighting, with $\tau$ as the fuzzy indices for weighting, where $\tau$ effectively controls the distributions of $w_{ik}$ in the clustering procedure; for example, $w_{ik} \to 1/D$ for a very large $\tau$. On the other hand, another category of ISSC algorithms has been developed based on the FCM/K-means model and entropy weighting. Unlike the fuzzy weighting based algorithms described in the previous subsection, the weighting in this category of algorithms is controllable by entropy. The review of ISSC algorithms presented here is organized as follows. First, ISSC algorithms based on the K-means framework, the FCM framework, and the Gaussian mixture model framework are described and analyzed respectively, followed by classic ISSC algorithms that have been developed based on other models. Some representative ISSC algorithms are listed in Table 7.

Table 7 Some Representative ISSC Algorithms

| Algorithm | Framework adopted for objective function | Method of feature weighting |
|---|---|---|
| *(1) AWA* | K-means | Fuzzy weighting |
| *(2) FWKM* | | Fuzzy weighting |
| *(3) FSC* | | Fuzzy weighting |
| *(4) EWKM* | | Entropy weighting |
| *(5) LAC* | | Entropy weighting |
| *(6) AWFCM* | FCM | Fuzzy weighting |
| *(7) SCAD* | | Fuzzy weighting |
| *(8) FPC/MPC* | Gaussian mixture model | Others |
| *(9) EWMM* | | Entropy weighting |
| *(10) COSA* | Others | Entropy weighting |

### 4.1 ISSC based on Multi-weighting K-means

Most ISSC algorithms are based on the classic K-means framework. The representative algorithms are discussed below.

*(1) AWA, FWKM, and FSC*

The Attribute Weighting Algorithm (AWA) developed by Chan et al. [11] employs an objective function similar to that of the W-k-means algorithm [39]. However, the shared weights $w_k$ of the $k$th feature of all of the clusters [39] are replaced by the weights $w_{ik}^\tau$ of the $k$th feature of each cluster. The objective function is expressed as,

$$\min J_{AWA}(\mathbf{U},\mathbf{V},\mathbf{W}) = \sum_{i=1}^{C}\sum_{j=1}^{N}u_{ij}\sum_{k=1}^{D}w_{ik}^{\tau}(x_{jk}-v_{ik})^2$$

s.t. $u_{ij} \in \{0,1\}$, $\sum_{i=1}^{C}u_{ij}=1$, $0 \le w_{ik} \le 1$, $\sum_{k=1}^{D}w_{ik}=1$.

AWA can effectively find the important features of each cluster. However, the weakness of AWA is that when some of the attributes have a standard deviation of zero, the algorithm will fail to work since the zeroes may be taken as the denominator in the learning rules. Improved versions have been proposed to overcome this weakness, including the Fuzzy Weighting K-Means (FWKM) algorithm [42] and the Fuzzy Subspace Clustering (FSC) algorithm [32, 33].

The objective function of FWKM is as follows:

$$\min J_{FWKM}(\mathbf{U},\mathbf{V},\mathbf{W}) = \sum_{i=1}^{C}\sum_{j=1}^{N}u_{ij}\sum_{k=1}^{D}w_{ik}^{\tau}\left[(x_{jk}-v_{ik})^2+\sigma\right]$$

$$\sigma = \frac{\sum_{j=1}^{N}\sum_{k=1}^{D}(x_{jk}-o_k)^2}{N \cdot D}, \quad o_k = \sum_{j=1}^{N}x_{jk}\Big/N$$

s.t. $u_{ij} \in \{0,1\}$, $\sum_{i=1}^{C}u_{ij}=1$, $0 \le w_{ik} \le 1$, $\sum_{k=1}^{D}w_{ik}=1$.

In the objective function of FWKM, a minor constant $\sigma$ is added when the distance is computed, which effectively avoids the issue in AWA caused by the possible standard deviations of zeroes in some attributes.

Gan et al. proposed the FSC algorithm [32, 33] using an objective similar to that of the FWKM algorithm [42] discussed previously. A detailed analysis of the properties of FSC can be found in [33].

$$\min J_{FSC}(\mathbf{U},\mathbf{V},\mathbf{W}) = \sum_{i=1}^{C}\sum_{j=1}^{N}u_{ij}\sum_{k=1}^{D}w_{ik}^{\tau}(x_{jk}-v_{ik})^2 + \varepsilon_0\sum_{i=1}^{C}\sum_{k=1}^{D}w_{ik}^{\tau}$$

s.t. $u_{ij} \in \{0,1\}$, $\sum_{i=1}^{C}u_{ij}=1$, $0 \le w_{ik} \le 1$, $\sum_{k=1}^{D}w_{ik}=1$.

FSC also introduces a constant to avoid the issue of zero standard deviations in AWA. Different from FWKM, the constant parameter $\varepsilon_0$ that is introduced in FSC needs to be set manually, while the constant parameter $\sigma$ in FWKM is set with a predefined formulation.

All of the three algorithms discussed above adopt the same feature weighting strategy, i.e., fuzzy weighting, where the fuzzy index of feature weights is an important parameter for controlling the distribution of the feature weights.

*(2) EWKM and LAC*

Besides fuzzy weighting, another important way to control the weight distribution is to adopt the maximum entropy strategy, i.e., entropy weighting. Representative entropy weighting ISSC algorithms include the Entropy Weighting K-Means (EWKM) clustering algorithm [41] and the Local Adaptive Clustering (LAC) algorithm [25].

Jing et al. proposed the EWKM clustering algorithm [41] using the following objective function:

$$\min J_{EWKM}(\mathbf{U},\mathbf{V},\mathbf{W}) = \sum_{i=1}^{C}\sum_{j=1}^{N} u_{ij} \sum_{k=1}^{D} w_{ik}(x_{jk}-v_{ik})^2 + \gamma \sum_{i=1}^{C}\sum_{k=1}^{D} w_{ik} \ln w_{ik}$$

$$\text{s.t.} \quad u_{ij} \in \{0,1\}, \ \sum_{i=1}^{C} u_{ij}=1, \ 0 \le w_{ik} \le 1, \ \sum_{k=1}^{D} w_{ik}=1,$$

where the second term in the objective function is the negative Shannon entropy and $\gamma$ is used to balance its influence on the clustering procedure. By introducing the entropy term, the weight that is obtained can be effectively controlled by entropy. For example, if $\gamma$ is very large, the features will be assigned equal values. EWKM has become a benchmarking ISSC algorithm and has been further extended to develop various XSSC algorithms, e.g., the ESSC algorithm [20]. While the fuzzy weighting can be interpreted using fuzzy mathematics, the entropy weighting used in EWKM can be easily explained using the theory of entropy in physics. Thus, the physical meaning of the entropy weighting is very clear. When entropy weighting is used, the parameter $\gamma$ that is involved has an important influence on the clustering result. How this parameter is set is an important research topic.

For the LAC algorithm [25], the objective function can be expressed as

$$\min J_{LAC}(\mathbf{U},\mathbf{V},\mathbf{W}) = \sum_{i=1}^{C}\sum_{k=1}^{D} w_{ik} X_{ik} + \gamma \sum_{i=1}^{C}\sum_{k=1}^{D} w_{ik} \ln w_{ik},$$

$$X_{ik} = \left( \sum_{j=1}^{N} u_{ij}(x_{jk}-v_{ik})^2 \right) \Big/ \sum_{j=1}^{N} u_{ij}$$

$$\text{s.t.} \quad u_{ij} \in \{0,1\}, \ \sum_{i=1}^{C} u_{ij}=1, \ 0 \le w_{ik} \le 1, \ \sum_{k=1}^{D} w_{ik}=1.$$

The objective functions of EWKM and LAC are indeed very similar. The only difference is that the effect of cluster size is considered in LAC, but disregarded in EWKM. The two algorithms LAC and EWKM essentially have similar advantages and disadvantages.

In addition to the entropy weighting LAC in [25], Domeniconi proposed an alternative LAC algorithm with a different objective function [26] as follows:

$$\max J_{O-LAC}(\mathbf{U},\mathbf{V},\mathbf{W}) = \sum_{i=1}^{C}\sum_{k=1}^{D} w_{ik} \exp(h \cdot X_{ik}),$$

$$X_{ik} = \left(\sum_{j=1}^{N} u_{ij}(x_{jk}-v_{ik})^2\right) \bigg/ \sum_{j=1}^{N} u_{ij}$$

s.t. $u_{ij} \in \{0,1\}$, $\sum_{i=1}^{C} u_{ij} = 1$, $0 \leq w_{ik} \leq 1$, $\sum_{k=1}^{D} w_{ik}^2 = 1$.

Note that the objective function is maximized to solve the solution variables, which is distinct from the entropy-based LAC approach [25]. In general, the experimental studies in the literature [20] show that the performance of the LAC algorithm in [26] is inferior to that of the LAC in [25].

### 4.2 ISSC based on Multi-weighting FCM

The algorithms discussed in the previous subsection are all based on the classic K-means framework. In order to be more adaptive to noisy data, improved versions of the algorithms have been developed based on the FCM framework. Representative ISSC algorithms using this approach are Attribute Weighting Fuzzy Clustering (AWFCM) and the Simultaneous Clustering and Attribute Discrimination (SCAD) methods.

*(1) AWFCM*

Keller and Klawonn proposed the AWFCM based on the FCM model with the objective function below [43]:

$$\min J_{AWFC}(\mathbf{U},\mathbf{V},\mathbf{W}) = \sum_{i=1}^{C}\sum_{j=1}^{N} u_{ij}^m \sum_{k=1}^{D} w_{ik}^\tau (x_{jk}-v_{ik})^2$$

s.t. $u_{ij} \in [0,1]$, $\sum_{i=1}^{C} u_{ij} = 1$, $0 \leq w_{ik} \leq 1$, $\sum_{k=1}^{D} w_{ik} = 1$.

From the objective function of AWFCM [11], it is clear that AWFCM is a soft partition clustering version of AWA, which makes it more adaptive to noisy data.

To our knowledge, AWFCM is the first fuzzy weighing ISSC algorithm. Based on AWFCM, improved versions of the algorithm have been proposed, e.g., the EFWSSC algorithm [46].

*(2) SCAD*

Frigui and Nasraoui proposed two versions of SCAD algorithms: SCAD-1 and SCAD-2. The following are the corresponding objective functions [29]:

$$\min J_{SCAD-1}(\mathbf{U},\mathbf{V},\mathbf{W}) = \sum_{i=1}^{C}\sum_{j=1}^{N}u_{ij}^2\sum_{k=1}^{D}w_{ik}(x_{jk}-v_{ik})^2 + \sum_{i=1}^{C}\delta_i\sum_{k=1}^{D}w_{ik}^2$$

s.t. $u_{ij}\in[0,1]$, $\sum_{i=1}^{C}u_{ij}=1$, $0\leq w_{ik}\leq 1$, $\sum_{k=1}^{D}w_{ik}=1$,

$$\min J_{SCAD-2}(\mathbf{U},\mathbf{V},\mathbf{W}) = \sum_{i=1}^{C}\sum_{j=1}^{N}u_{ij}^m\sum_{k=1}^{D}w_{ik}^{\tau}(x_{jk}-v_{ik})^2$$

s.t. $u_{ij}\in[0,1]$, $\sum_{i=1}^{C}u_{ij}=1$, $0\leq w_{ik}\leq 1$, $\sum_{k=1}^{D}w_{ik}=1$.

Comparing the objective functions of AWFCM and SCAD-2, it is evident that they are essentially fuzzy weighting SSC algorithms of the same kind. Thus, SCAD-2 has the same characteristics as AWFCM. Comparing SCAD-2 with SCAD-1, it is obvious that the former is a more general approach than the latter.

### 4.3 Probability Mixture Model based ISSC

Although the ISSC algorithms based on the FCM/K-means framework have demonstrated promising performance in different applications, these algorithms do not take into consideration the effect of data distribution on the clustering procedure. Thus, the probability mixture model has been adopted to improve the ISSC algorithms to address this issue. In this subsection, two representative algorithms are introduced.

*(1) FPC/MPC*

Chen et al. proposed the Fuzzy/Model Projective Clustering (FPC/MPC) algorithm based on the mixture model [13,14]. In FPC/MPC, each projected dimension is assumed to fit in the Gaussian mixture distribution as follows:

$$F(x_k;\Theta_k) = \sum_{i=1}^{C}\alpha_i G(x_k|v_{ik},\sigma_i)$$

$$G(x_k|v_{ik},\sigma_i) = \frac{1}{\sqrt{2\pi}\sigma_k}\exp\left(-\frac{w_{ik}}{2\sigma_i^2}\left((\mathbf{x}-\mathbf{v}_i)\cdot\mathbf{e}_{ik}^T\right)^2\right)$$

with $\sum_{k=1}^{D} \sqrt{w_{ik}} = 1$, where $v_{ik}, \sigma_i$ denote the means and variances, respectively. Furthermore, the following Kullback-Leibler (KL) divergence is minimized for parameter learning:

$$R(\widehat{\Theta}_j) = \int F(x_k; \Theta_j) \ln \frac{F(x_k; \Theta_j)}{\widehat{F}(x_k; \widehat{\Theta}_j)} dx_j.$$

Based on the above criterion, the objective function for clustering is finally given by

$$J_{FPC/MPC}(\mathbf{U}, \mathbf{V}, \mathbf{W}, \boldsymbol{\alpha}, \boldsymbol{\sigma}, \mathbf{e}) = \sum_{i=1}^{C} \sum_{j=1}^{N} \sum_{k=1}^{D} \frac{u_{ij} w_{ik}}{2\sigma_i^2} \left((\mathbf{x}_j - \mathbf{v}_i) \cdot \mathbf{e}_{ik}^T\right)^2 - \sum_{i=1}^{C} D \ln \frac{\alpha_i}{\sqrt{2\pi}\sigma_i} \sum_{j=1}^{N} u_{ij} + D \sum_{i=1}^{C} \sum_{j=1}^{N} u_{ij} \ln u_{ij}$$

s.t. $u_{ij} \in \{0,1\}$, $\sum_{i=1}^{C} u_{ij} = 1$, $0 \leq w_{ik} \leq 1$, $\sum_{k=1}^{D} \sqrt{w_{ik}} = 1$, $0 \leq \alpha_i \leq 1$, $\sum_{i=1}^{C} \alpha_i = 1$.

When the axis-aligned subspace is only considered, the above objective function can be reduced to

$$J_{FPC/MPC}(\mathbf{U}, \mathbf{V}, \mathbf{W}, \boldsymbol{\alpha}, \boldsymbol{\sigma}) = \sum_{i=1}^{C} \sum_{j=1}^{N} \sum_{k=1}^{D} \frac{u_{ij} w_{ik}}{2\sigma_i^2} (x_{jk} - v_{ik})^2 - \sum_{i=1}^{C} D \ln \frac{\alpha_k}{\sqrt{2\pi}\sigma_i} \sum_{j=1}^{N} u_{ij} + D \sum_{i=1}^{C} \sum_{j=1}^{N} u_{ij} \ln u_{ij}.$$

It can be seen that the FPC/MPC algorithm also contains an entropy term for the clustering procedure; however, the entropy term here is used to control the partition $u_{ij}$ instead of the feature weight $w_{ij}$. When compared with the classic ISSC algorithms, which are based on the FCM/K-means model, the FPC algorithm, which is based on the mixture model, is expected to possess a stronger ability to adapt to data distributions, as is evident from the promising clustering results achieved on high-dimensional data [54]. However, the ISSC algorithms based on the mixture model are much more complicated than the FCM/K-means based algorithms, and are therefore not very popular in the field of SSC. Thus, when the Gaussian distribution (or any other distribution) is applicable, it makes sense to take it into consideration in the clustering process. Otherwise, a non-probabilistic tool should be employed instead, as discussed in [38].

*(2) EWMM*

Peng and Zhang proposed the entropy weighting mixture model (EWMM) algorithm [54]. By using an approach similar to that in FPC/MPC, the objective function of EWMM can be formulated as follows:

$$J(\mathbf{U},\mathbf{V},\mathbf{W},\boldsymbol{\alpha},\boldsymbol{\sigma}) = \frac{1}{N}\sum_{i=1}^{C}\sum_{j=1}^{N}u_{ij}\left[-\ln\alpha_i + \sum_{k=1}^{D}\left(\frac{w_{ik}}{2\sigma_i^2}(x_{jk}-v_{ik})^2 - \ln\sqrt{\frac{w_{ik}}{2\sigma_i^2}}\right) + \ln u_{ij}\right]$$

s.t. $u_{ij} \in \{0,1\}$, $\sum_{i=1}^{C}u_{ij}=1$, $0 \leq w_{ik} \leq 1$, $\sum_{k=1}^{D}w_{ik}=1$, $0 \leq \alpha_i \leq 1$, $\sum_{i=1}^{C}\alpha_i = 1$.

Furthermore, with the extra control over weighting obtained by using entropy, the final objective function for clustering is given by

$$J_{EWMM}(\mathbf{U},\mathbf{V},\mathbf{W},\boldsymbol{\alpha},\boldsymbol{\sigma}) = \frac{1}{N}\sum_{i=1}^{C}\sum_{j=1}^{N}u_{ij}\left[-\ln\alpha_i + \sum_{k=1}^{D}\left(\frac{w_{ik}}{2\sigma_i^2}(x_{jk}-v_{ik})^2 - \ln\sqrt{\frac{w_{ik}}{2\sigma_i^2}}\right) + \ln u_{ij}\right] + \sum_{i=1}^{C}\lambda_i\sum_{k=1}^{D}w_{ik}\ln w_{ik}.$$

Note that although both algorithms based on mixture models (i.e., the Fuzzy/Model algorithm and the EWMM algorithm) contain an entropy term, these terms have very different purposes. The former uses the entropy term to control the partition $u_{ij}$, while the latter uses it to control the weight $w_{ik}$. As an FPC algorithm, EWMM is able to adapt to data distributions better. However, this algorithm is not popular, since it is more complicated than FCM/K-means based algorithms.

### 4.4 ISSC based on Other Models

In addition to the ISSC algorithms based on FCM/K-Means and the mixture model, other ISSC algorithms have also been proposed. Among them, the Clustering Objects on Subsets of Attributes (COSA) proposed by Friedman and Meulman [28] is a representative algorithm that uses the following objective function:

$$J_{COSA}(\mathbf{U},\mathbf{W}) = \sum_{i=1}^{C}\frac{\Lambda_i}{n_i^2}\sum_{z(j)=z(j')=i}\left[\sum_{k=1}^{D}(w_{ik}d_{jj'k} + \alpha w_{ik}\ln w_{ik}) + \alpha\ln(D)\right]$$

$$= \sum_{i=1}^{C}\frac{\Lambda_i}{n_i^2}\sum_{k=1}^{D}\left(w_{ik}\sum_{z(j)=z(j')=i}d_{jj'k}\right) + \alpha\sum_{i=1}^{C}\frac{\Lambda_i}{n_i^2}\sum_{k=1}^{D}w_{ik}\ln w_{ik} + \alpha\sum_{i=1}^{C}\frac{\Lambda_i}{n_i^2}\sum_{k=1}^{D}\ln(D)$$

s.t. $0 \leq w_{ik} \leq 1$, $\sum_{k=1}^{D}w_{ik}=1$,

where $\Lambda_i$ are the cluster weights; $n_i$ is the number of objects assigned to the $i$th cluster; $z(j)$ is an encode function to map each object $j$ to a special group $G_l$ $(1 \leq l \leq C)$; and $d_{jj'k}$ denotes the distances between pairs of objects $(j, j')$ assigned to the same group. COSA is an entropy-based subspace clustering algorithm. As discussed in [26], one shortcoming of COSA is

### 4.5 Simulation on the Toy Dataset

In this section, as one of the classic ISSC algorithms, FSC has been adopted to validate the performance of ISSC algorithms on Toy-D. The performance of FSC is given in Fig. 3 and in Tables 8 and 9. From Fig. 3 and Table 8, we can see that FSC can effectively assign the important features with the larger values of weights for each cluster. Table 8 shows that the FSC algorithm effectively found all the important features of the second and third clusters in Toy-D. Meanwhile, some important features of the first cluster were also obtained. Thus, as an ISSC algorithm, FSC has demonstrated a better ability to find the important features of clusters than the classic CSSC algorithm W-$k$-means. Table 9 shows that FSC has better clustering performance than W-$k$-means on Toy-D.

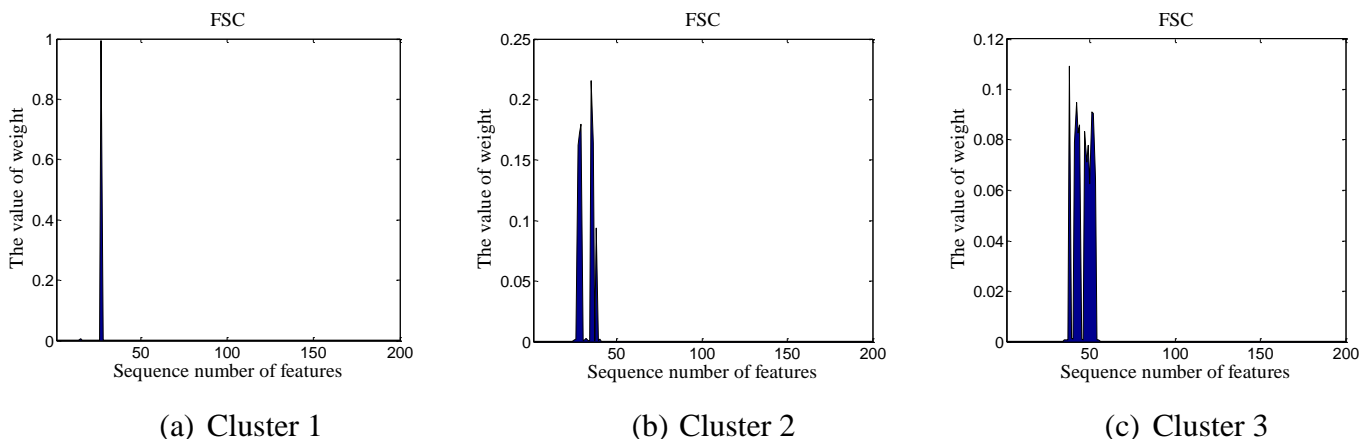

(a) Cluster 1      (b) Cluster 2      (c) Cluster 3

**Fig. 3 The distributions of the weights of each cluster obtained by the FSC algorithm on Toy-D.**

**Table 8 Comparison of the feature identification abilities of FSC on Toy-D.**

| No. of clusters | True | Identified by FSC |
|---|---|---|
| 1 | 30 features: {1:30} | 30 features with top weights: {**27,15,29,14,6,11,20,13,26,4,28,8,17**,31,**5,24,16,21,25, 23,30**,35,39,47,41,**9,3,19**,32,**1**} |
| 2 | 26 features: {20:45} | 26 features with top weights: {**35,29,28,36,27,38,32,40,26,31,25,37,39,34,30,33**,49,62, 80,129,89,127,190,150,186,120} |
| 3 | 21 features: {35:55} | 21 features with top weights: {**38,42,51,52,44,47,43,41,49,48,53,50,39,45,40,35, 55,36,54,37,46**} |
| Identification Rate | | (24+16+21)/77=61/77 |

The numbers in boldface denote the truly important features that were detected as being associated with the embedded subspaces.

Table 9 Comparison of the clustering performance of W-*k*-means and FSC on Toy-D.

| Index | | W-*k*-means | FSC |
|---|---|---|---|
| *RI* | *mean* | 0.9708 | 0.9729 |
| | *std* | 0.0117 | 0.0253 |
| | * | 0.9842 | 1 |
| *NMI* | *mean* | 0.8727 | 0.9109 |
| | *std* | 0.0327 | 0.0771 |
| | * | 0.9080 | 1 |

* Denotes the values of RI and NMI achieved by each algorithm when the lowest value of the loss function is obtained within the 10 runs.

## 5 XSSC

XSSC algorithms refer to a category of SSC methods that have been developed to improve the clustering performance of CSSC and ISSC algorithms by introducing new learning mechanisms for specific purposes. Table 10 lists seven subcategories of the XSSC algorithm. The eighth subcategory, i.e., others, is not included in the table. Some XSSC algorithms have been developed for general purposes, including XSSC algorithms that integrate between-clustering information, evolutionary optimization learning, adaptive metrics, and ensemble learning, respectively. On the other hand, XSSC algorithms have also been developed for specific purposes, e.g., multi-view data or data with imbalanced clusters. The representative XSSC algorithms are reviewed in this subsection.

Table10 Some Representative XSSC Algorithms

| Algorithm | Mechanism adopted | Purpose |
|---|---|---|
| *(1) IEWKM* | Between-cluster separation integrated | General purpose |
| *(2) ESSC* | | |
| *(3) EFWSSC* | | |
| *(4) Coevolutionary SSC* | Evolutionary learning | General purpose |
| *(5) PSOVW* | | |
| *(6) MOEA-SSC* | | |
| *(7) MWK-Means* | Adaptive metric | General purpose |
| *(8) WFKCA* | | |
| *(9) EM-PCE* | Ensemble learning | General purpose |
| *(10) MOEA-PCE* | | |
| *(11) TW-k-means* | Multi-view learning | For multi-view data |
| *(12) FG-k-means* | | |
| *(13) WLAC* | Imbalanced data learning | For data with imbalanced clusters |
| *(14) RKM* | Transformed feature space | General purpose |
| *(15) FKM* | | |

### 5.1 Between-cluster Separation

Most SSC algorithms perform clustering by optimizing the within-cluster compactness without making use of the between-cluster information. Recently, algorithms integrating between-cluster separation with within-cluster compactness have been developed to enhance the clustering performance. Three representative algorithms are the Improved Entropy Weighting K-means (IEWKM) algorithm [46], the Enhanced SSC (ESSC) algorithm [20], and the Enhanced Fuzzy Weighting Soft Subspace Clustering (EFWSSC) algorithm [36].

*(1) IEWKM*

By integrating the between-cluster separation, Li et al. proposed the Improved Entropy IEWKM with the following objective function [46]:

$$J_{IEWKM}(\mathbf{U},\mathbf{V},\mathbf{W}) = \frac{\sum_{i=1}^{C}\sum_{j=1}^{N}u_{ij}\sum_{k=1}^{D}w_{ik}(x_{jk}-v_{ik})^2}{\sum_{i=1}^{C}\sum_{k=1}^{D}w_{ik}(v_{ik}-v_{0k})^2} + \gamma\sum_{i=1}^{C}\sum_{k=1}^{D}w_{ik}\ln w_{ik}$$

s.t. $u_{ij}\in\{0,1\}$, $\sum_{i=1}^{C}u_{ij}=1$, $0\le w_{ik}\le 1$, $\sum_{k=1}^{D}w_{ik}=1$.

Although IEWKM effectively uses the between-cluster information, one weakness of this algorithm is that it is not easy to optimize the objective function. In addition, the mathematical derivation of the learning rules for the cluster centers in this algorithm lacks rigorousness [36].

*(2) ESSC and EFWSSC*

Besides IEWKM, ESSC and EFWSSC are two other algorithms that integrate between-cluster separation. The derivation of these two algorithms is mathematically more rigorous. The ESSC algorithm [20], which is based on the EWKM method [41], has been proposed to improve clustering performance by simultaneously minimizing the within-cluster compactness in the weighting subspace and maximizing the between-cluster separation. The objective function of the ESSC algorithm is given by

$$J_{ESSC}(\mathbf{U},\mathbf{V},\mathbf{W}) = \sum_{i=1}^{C}\sum_{j=1}^{N}u_{ij}^{m}\sum_{k=1}^{D}w_{ik}(x_{jk}-v_{ik})^2 + \gamma\sum_{i=1}^{C}\sum_{k=1}^{D}w_{ik}\ln w_{ik} - \eta\sum_{i=1}^{C}(\sum_{j=1}^{N}u_{ij}^{m})\sum_{k=1}^{D}w_{ik}(v_{ik}-v_{0k})^2$$

s.t. $0\le u_{ij}\le 1$, $\sum_{i=1}^{C}u_{ij}=1$, $0\le w_{ik}\le 1$, $\sum_{k=1}^{D}w_{ik}=1$.

The objective function contains three terms: the weighting within-cluster compactness, the entropy of weights, and the weighting between-cluster separation. The first and second terms are directly inherited from the objective function of EWKM subspace clustering, except that the k-means model is replaced by the FCM model. In this objective function, the parameters $\gamma$ ($\gamma > 0$) and $\eta$ ($\eta > 0$) are used to control the influence of entropy and the weighting between-cluster separation, respectively. It is worth pointing out that when $m \to 1$ and $\eta = 0$, the ESSC algorithm is reduced to the EWKM algorithm [26]. Thus, the EWKM algorithm can be regarded as a special case of the ESSC algorithm. ESSC has demonstrated distinct advantages compared to the classic ISSC algorithms, as reported in [20]. However, one disadvantage of ESSC is that more parameters need to be set manually and the setting of appropriate parameters is still an open problem in ESSC.

Using a similar strategy, Guan improved the FSC algorithm [32,33] by making use of between-cluster separation and proposing the EFWSSC algorithm [36]. The objective function of the EFWSSC algorithm is

$$J_{EFWSSC}(\mathbf{U},\mathbf{V},\mathbf{W}) = \sum_{i=1}^{C}\sum_{j=1}^{N} u_{ij}^m \sum_{k=1}^{D} w_{ik}^{\tau}(x_{jk}-v_{ik})^2 + \varepsilon \sum_{i=1}^{C}\sum_{k=1}^{D} w_{ik}^{\tau} - \eta \sum_{i=1}^{C}(\sum_{j=1}^{N} u_{ij}^m)\sum_{k=1}^{D} w_{ik}^{\tau}(v_{ik}-v_{0k})^2$$

s.t. $0 \leq u_{ij} \leq 1$, $\sum_{i=1}^{C} u_{ij} = 1$, $0 \leq w_{ik} \leq 1$, $\sum_{k=1}^{D} w_{ik} = 1$.

Similarly, when $m \to 1$ and $\eta = 0$, the EFWSSC algorithm degenerates into the FSC algorithm, i.e., the FSC algorithm is a special case of the EFWSSC algorithm. EFWSSC is very similar to ESSC. The only difference is that entropy weighting and fuzzy weighting are used by ESSC and EFWSSC, respectively. Thus, they have similar advantages and disadvantages.

### 5.2 Evolutionary Learning

Most existing SSC algorithms rely on an iterative learning strategy to optimize the objective functions using a method similar to the classic K-means and FCM methods. However, these algorithms suffer from poor initialization sensitivity and local optimization. In order to overcome such deficiencies, an evolutionary learning technique has been introduced to optimize the objective functions of SSC. Several representative XSSC algorithms that were developed based on evolutionary learning are reviewed below.

*(1) Coevolutionary SSC*

Gangrski et al. proposed two SSC algorithms based on coevolutionary learning [34]. The first algorithm was inspired by the Lamarck theory and used the distance-based cost function defined in the AWC algorithm [11] as the fitness function. The second algorithm employed a fitness function based on a new partitioning quality measure. The experimental results in [34] highlighted the benefits of using coevolutionary feature weighting methods to improve the knowledge discovery process. The experimental results showed that these algorithms come close to outperforming the K-means-like algorithms. However, their shortcoming is that they require more CPU time than algorithms based on hill-climbing optimization techniques.

*(2) PSOVW*

Lu et al. proposed the Particle Swarm Optimizer for Variable Weighting (PSOVW) algorithm by using the particle swarm optimizer as the evolutionary strategy for SSC. The following objective function is employed for optimizing variables [49]:

$$J_{PSOVW}(\mathbf{U},\mathbf{V},\mathbf{W}) = \sum_{i=1}^{C}\sum_{j=1}^{N} u_{ij} \sum_{k=1}^{D} \frac{w_{ik}^{\tau}}{\sum_{k'=1}^{D} w_{ik'}^{\tau}} (x_{jk} - v_{ik})^2$$

s.t. $u_{ij} \in \{0,1\}$, $\sum_{i=1}^{C} u_{ij} = 1$, $0 \leq w_{ik} \leq 1$, $\sum_{k=1}^{D} w_{ik} = 1$.

By transforming the original constrained variable weighting problem into a problem with bound constraints, i.e., using a normalized representation of variable weights, the particle swarm optimizer can easily minimize the objective function to search for global optima for the variable weighting problem. The experimental results show that the PSOVW algorithm greatly improves the clustering performance and that the clustering results are much less dependent on the initial cluster centers. Although PSOVW overcomes the issue of the initial sensitivity of methods based on the classic KM/FCM framework, some new parameters of the PSO optimizer, such as the size of the population, need to be set manually. In addition, since PSO is a kind of random optimization algorithm, this causes the solutions to be unstable in different runs.

*(3) MOEA-SSC*

While the two SSC algorithms based on evolutionary learning discussed above both use a single objective function for optimization, algorithms based on multi-objective evolutionary learning have also been investigated. One representative algorithm is the Multi-Objective

Evolutionary Approach for SSC (MOEA-SSC) developed by Xia et al. [72], which employs new encoding and operators. Here, two objective functions are adopted for SSC, i.e.,

$$J_{IN}(\mathbf{U},\mathbf{V},\mathbf{W}) = \sum_{i=1}^{C}\sum_{j=1}^{N} u_{ij}^{m} \sum_{k=1}^{D} w_{ik}(x_{jk}-v_{ik})^2$$

$$J_{Add}(\mathbf{U},\mathbf{V},\mathbf{W}) = \sum_{i=1}^{C} \frac{A_i}{\sum_{i'=1}^{C}(v_{ik}-v_{i'k})^2} + \sum_{i=1}^{C}\sum_{k=1}^{D} w_{ik} \ln w_{ik}$$

with $A_i = \dfrac{\sum_{k=1}^{D} w_{ik}\delta_k}{\sum_{k=1}^{D}\delta_k}$, $\delta_k = \begin{cases} 1 & \text{if } w_{ik} > 1/D \\ 0 & \text{Otherwise} \end{cases}$

s.t. $0 \leq u_{ij} \leq 1$, $\sum_{i=1}^{C} u_{ij}=1$, $0 \leq w_{ik} \leq 1$, $\sum_{k=1}^{D} w_{ik}=1$,

where $J_{IN}$ is the within-cluster dispersion and $J_{Add}$ contains the information on both the negative weight entropy and the separation between clusters. Similar to other SSC algorithms based on evolutionary learning, the MOEA-SSC algorithm is less dependent on the initial cluster centers. Like other algorithms based on evolutionary learning, MOEA-SSC is less sensitive to initialization. However, MOEA-SSC is also a random optimization algorithm and the stability of the solutions in different runs cannot be guaranteed.

### 5.3 Adaptive Metric

In SSC, Euclidean distance is commonly used as the metric for defining the distance between the data points and the cluster centers in each dimension. In order to improve clustering performance, a number of modified metrics have been proposed [19,58,63]. Two of the metrics are introduced below, along with the corresponding XSSC algorithms that were developed.

*(1) Minkowski Metric*

Amorim and Mirkin attempted to improve fuzzy weighting SSC algorithms by employing an alternative metric, i.e., the Minkowski metric, to replace the common Euclidean distance. The algorithm that was developed is known as the Minkowski metric Weighted K-Means (MWK-Means) algorithm [19], and the objective function of the algorithm is

$$J_{MWK-Means}(\mathbf{U},\mathbf{V},\mathbf{W}) = \sum_{i=1}^{C}\sum_{j=1}^{N} u_{ij} \sum_{k=1}^{D} w_{ik}^{\tau} |x_{jk} - v_{ik}|^{p},$$

s.t. $u_{ij} \in \{0,1\}$, $\sum_{i=1}^{C} u_{ij} = 1$, $0 \leq w_{ik} \leq 1$, $\sum_{k=1}^{D} w_{ik} = 1$,

where the traditional Euclidean distance is replaced by the Minkowski metric with $p$ as the parameter. By comparing MWK-Means with AWA [11], we can easily see that when $p=2$, AWA is indeed a special case of MWK-Means. For MWK-Means, an important task is to determine the appropriate value for $p$. The optimal setting of $p$ depends on the given clustering task.

*(2) Kernel Metric*

Besides the Minkowski Metric, a more adaptive metric, i.e., the kernel metric, has been adopted for SSC. Shen et al. developed an improved version of the fuzzy weighting SSC algorithm, namely, the Weighted Fuzzy Kernel-Clustering Algorithm (WFKCA), by proposing a new metric in the Mercer kernel space, i.e., Mercer kernel distance [58]. This metric is used instead of the Euclidean distance in the original space for each dimension. The objective function of the WFKCA can be formulated as follows:

$$J_{WFKCA}(\mathbf{U},\mathbf{V},\mathbf{W}) = \sum_{i=1}^{C}\sum_{j=1}^{N} u_{ij}^{m} \sum_{k=1}^{D} w_{ik}^{\tau} \|\varphi(x_{jk}) - \varphi(v_{ik})\|^{2}$$

$$\|\varphi(x_{jk}) - \varphi(v_{ik})\|^{2} = k(x_{jk}, x_{jk}) + k(x_{jk}, x_{jk}) - 2k(x_{jk}, x_{jk})$$

s.t. $0 \leq u_{ij} \leq 1$, $\sum_{i=1}^{C} u_{ij} = 1$, $0 \leq w_{ik} \leq 1$, $\sum_{k=1}^{D} w_{ik} = 1$,

where $\varphi(x_{jk})$ is the mapped feature vector of feature $x_{jk}$ and $k(x_{jk}, x_{jk})$ is the kernel function. With Mercer kernel distance as the metric, the WFKCA can improve the clustering performance and adaptive abilities of the algorithm, which are better than those of algorithms based on Euclidean distance. In general, the SSC algorithm based on kernel metric has a better ability to adapt to different data distributions than traditional SSC algorithms based on Euclidian metric. A critical challenge for the SSC algorithms based on kernel metric is the selection of the appropriate kernel function and the setting of the corresponding kernel parameters. This problem has yet to be resolved, and further study is required.

## 5.4 Ensemble Learning

Many SSC algorithms are parameter sensitive, i.e., the clustering performance is severely influenced by the parameter setting. To tackle this issue and improve clustering performance, ensemble learning is a useful technique that involves fusing the clustering results obtained with different parameter settings on the same dataset. Domeniconi and Al-razgan proposed the following two clustering ensemble approaches based on graph partitioning to overcome the problem of parameter insensitivity in the LAC algorithm: the Weighted Similarity Partitioning Algorithm (WSPA) and the Weighted Bipartite Partitioning Algorithm (WBPA) [24]. The WSPA combines multiple clustering results obtained from different runs of the LAC SSC algorithm. Studies have shown that the WSPA is highly effective [20,24]. One disadvantage of the WSPA is the high level of computational complexity involved. The amount of clustering results obtained under different parameter settings, which should be used for ensemble learning, is also an open problem.

Gullo et al. also proposed Projective Clustering Ensemble (PCE) algorithms for the ensemble learning of soft clustering [37]. The PCE algorithm was developed with two different formulations, namely, single-objective PCE and two-objective PCE. The former is implemented as an expectation maximization (EM) like algorithm, and is thus called EM-PCE. The latter employs techniques similar to those in the realm of multi-objective evolutionary algorithms, and is known as the MOEA-PCE algorithm. An experimental evaluation shows that MOEA-PCE generally produces higher-quality projective consensus clustering results than EM-PCE. However, the EM-PCE algorithm is more efficient than the EM-PCE algorithm since the computational complexity of the multi-objective evolutionary based MOEA-PCE is quite high.

## 5.5 Multi-view Learning

As traditional SSC algorithms are not suitable for multi-view data, multi-view learning has been proposed [12,16,30] and is becoming a popular approach in the field of machine learning. Two representative SSC algorithms based on multi-view learning are introduced below.

*(1) TW-k-means*

Chen et al. proposed the Two-level variable Weighting k-means (TW-k-means) algorithm by introducing the view weighting mechanism [12]. This is a multi-view SSC algorithm that is based on traditional weighting clustering methods. The objective function of TW-k-means is given by

$$J_{EWKM}(\mathbf{U},\mathbf{V},\mathbf{w},\tilde{\mathbf{w}}) = \sum_{i=1}^{C}\sum_{j=1}^{N}\sum_{t=1}^{T}\sum_{k\in G_t} u_{ij}\tilde{w}_t w_k (x_{jk}-v_{ik})^2 + \gamma_1 \sum_{t=1}^{T}\sum_{k\in G_t} w_k \ln w_k + \gamma_2 \sum_{t=1}^{T} \tilde{w}_t \ln \tilde{w}_t$$

s.t. $u_{ij} \in \{0,1\}$, $\sum_{i=1}^{C} u_{ij}=1$, $0\leq w_k \leq 1$, $\sum_{k\in G_t} w_k =1$, $t=1,\cdots,T$, $0\leq \tilde{w}_t \leq 1$, $\sum_{t\in 1}^{T}\tilde{w}_t =1$,

where $T$ is the total number of views; $G_t$ is the set of the dimensions in the $t$th view ($1\leq t \leq T$); $\tilde{w}_t$ is the weight of the $t$th view; and $w_k$ is the weight of the $k$th dimensional in the $t$th view.

In this algorithm, feature weighting is performed in the first level weighting, where the importance of the features in each view is adjusted. View weighting is then performed in the second level weighting, where the influence of each view in the clustering procedure is adjusted. TW-k-means effectively realizes the co-learning of different views. However, the co-learning mechanism is still very simple and only the objective functions of different views are adjusted adaptively. More advanced co-learning mechanisms, such as the co-learning of the fuzzy partition matrices of different views, are expected to further improve the clustering performance.

*(2) FG-k-means*

TW-k-means remains a multi-view version of the CSSC algorithm, i.e., all of the clusters have a common feature weight vector. To extend TW-k-means for multi-feature weighting, the Feature Groups weighting K-means (FG-k-means) algorithm is introduced [16]. The FG-k-means algorithm is based on the EWKM algorithm, and the corresponding objective function is given by

$$J_{EWKM}(\mathbf{U},\mathbf{V},\mathbf{W},\tilde{\mathbf{w}}) = \sum_{i=1}^{C}\sum_{j=1}^{N}\sum_{t=1}^{T}\sum_{k\in G_t} u_{ij}\tilde{w}_t w_{ik}(x_{jk}-v_{ik})^2 + \gamma_1 \sum_{t=1}^{T}\sum_{k\in G_t} w_{ik} \ln w_{ik} + \gamma_2 \sum_{t=1}^{T} \tilde{w}_t \ln \tilde{w}_t$$

s.t. $u_{ij} \in \{0,1\}$, $\sum_{i=1}^{C} u_{ij}=1$, $0\leq w_{ik} \leq 1$, $\sum_{k\in G_t} w_{ik} =1$, $t=1,\cdots,T$, $0\leq \tilde{w}_t \leq 1$, $\sum_{t\in 1}^{T}\tilde{w}_t =1$,

where $T$ is the total number of views; $G_t$ is the set of dimensions in the $t$th view ($1\leq t \leq T$); $\tilde{w}_t$ is the weight of the $t$th view, and $w_{ik}$ is the weight of the $k$th dimension for the $i$th cluster in the $t$th view.

Comparing the FG-k-means and TW-k-means algorithms, it can be seen that the former is indeed an extension of the latter, from CSSC to ISSC. Like TW-k-means, the co-learning mechanism of FG-k-means for different views is simple and only the objective functions of different views are adjusted adaptively. To develop more effective multi-view SSC algorithms, advanced co-learning mechanisms are required.

## 5.6 Imbalanced Clusters

Datasets with imbalanced clusters are common in practical applications. The classic SSC algorithms cannot handle these datasets effectively and the performance of these algorithms will degrade. To solve this issue, an XSSC algorithm, called the Weighted LAC (WLAC), is proposed based on an elite selection of weighted clusters [53]. The objective function of WLAC can be expressed as follows:

$$J_{WLAC}(\mathbf{U},\mathbf{V},\mathbf{W},\mathbf{d}) = \sum_{i=1}^{C} d_i \sum_{k=1}^{D} w_{ik} X_{ik} + \gamma_1 \sum_{i=1}^{C} \sum_{k=1}^{D} w_{ik} \ln w_{ik} + \gamma_2 \sum_{i=1}^{C} d_i \ln d_i ,$$

$$X_{ik} = \left( \sum_{j=1}^{N} u_{ij} (x_{jk} - v_{ik})^2 \right) \bigg/ \sum_{j=1}^{N} u_{ij}$$

$$\text{s.t. } u_{ij} \in \{0,1\}, \ \sum_{i=1}^{C} u_{ij} = 1, \ 0 \leq w_{ik} \leq 1, \ \sum_{k=1}^{D} w_{ik} = 1, \ 0 \leq d_i \leq 1, \ \sum_{i=1}^{C} d_i = 1,$$

where $d_i$ is the coefficient representing the diameter of the $i$th cluster and is used to balance the influence of the imbalanced clusters. In particular, strategies such as ensemble learning have also been studied to reduce the influence of the parameters $\gamma_1$ and $\gamma_2$ in the WLAC algorithm. In WLAC, $\gamma_2$ is a parameter that is critical to controlling the influence of the different clusters. The setting of appropriate values for this parameter remains a challenging question.

## 5.7 Subspace Extraction in the Transformed Feature Space

Almost all of the SSC algorithms that were reviewed above extract subspace from the original full space. Alternative algorithms have been developed to obtain subspace from transformed feature space, such as algorithms based on principal component analysis (PCA) technology. Thus far, PCA has been used to build a low-dimensional subspace to make it easier for the clustering model to obtain the best partitioning of objects. Two representative PCA based *K*-means subspace clustering methods, Reduced *K*-means analysis (RKM) and Factorial *K*-means analysis (FKM), are introduced here [23,60]. The objective functions of these two methods are as follows:

$$F_{RKM}(\mathbf{U},\mathbf{V},\mathbf{A}) = \|\mathbf{X} - \mathbf{U}\mathbf{V}\mathbf{A}^T\|^2$$

and

$$F_{FKM}(\mathbf{U},\mathbf{V},\mathbf{A}) = \|\mathbf{X}\mathbf{A} - \mathbf{U}\mathbf{V}\|^2,$$

where $\mathbf{U}$ is a binary $N \times C$ membership matrix; $\mathbf{V}$ is a $C \times Q$ center matrix on the $Q$-dimensional subspace ($Q<D$); $\mathbf{A}$ is a $D \times Q$ columnwise orthonormal mapping matrix; and $\mathbf{A}^T\mathbf{A} = \mathbf{I}$. When $Q=D$, both RKM and FKM are reduced to the classic $K$-means method.

Timmerman et al. further analyzed these two methods and concluded that RKM generally performs better than FKM when the majority of the features reflect the clustering structure and/or when the features are standardized before analysis.

### 5.8 Other XSSC Algorithms

In addition to the seven broad categories of XSSC algorithms discussed above, many other XSSC algorithms have recently been reported [4,6,8,15,27,30,31,48,63,67]. A brief summary is given below.

*(1) Categorical Data*

Most SSC algorithms are only applicable to numeric data. The IEWKM algorithm has been proposed for a mixture of categorical and numeric data [48]. A modified version of EWKM [4] was also proposed by Ahmad and Dey for handling a mixture of categorical and numeric data. In addition, Bai et al. proposed a modified SSC algorithm based on AWA and the use of an improved metric for categorical data [6]. In [15], a soft subspace clustering algorithm with probabilistic distance is proposed for categorical data. Conducting a convergence analysis of these algorithms remains a challenging task.

*(2) Data Reliability*

Boongoen et al. proposed a novel approach called reliability-based SSC, based on measuring the reliability of the data [8]. This approach is advantageous in that it can be applied to various clustering algorithms, while existing wrappers are only suitable for the FCM/K-means model and/or a mixture model. Research has been conducted to increase the efficiency of reliability-based SSC algorithms and make them feasible for handling large datasets. An evaluation of gene expression data shows that the algorithms can improve the corresponding baseline techniques and outperform important soft and crisp subspace clustering methods.

### 5.9 Simulation on the Toy Dataset

In this section, ESSC, as a representative XSSC algorithm, is adopted to evaluate the performance of XSSC algorithms on Toy-D. The performance of ESSC is given in Fig. 4 and in Tables 11 and 12. In Fig. 4 and Table 11, we can see that ESSC has effectively assigned the important features with the larger values of weights for each cluster. In particular, Table 11 shows that ESSC is better than the classic ISSC algorithm FSC in identifying the important feature space for the Toy-D dataset. Table 12 also shows that the clustering performance achieved by ESSC on the Toy-D dataset is better than that achieved by W-k-means and FSC.

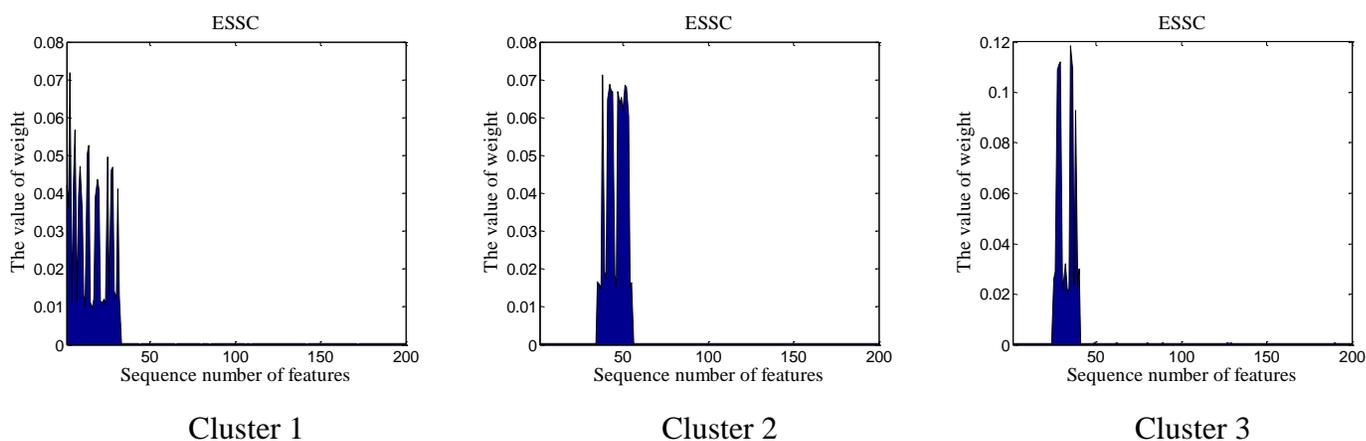

Cluster 1                Cluster 2                Cluster 3

**Fig. 4 The distributions of weights of each cluster obtained by the ESSC algorithm on Toy-D.**

**Table. 11 Comparison of the subspace identification abilities of FSC and ESSC on Toy-D.**

| No. of clusters | Sequence no. of important features | | |
|---|---|---|---|
| | True | Identified by FSC | Identified by ESSC |
| 1 | 30 features: {1:30} | 30 features with top weights: {**27,15,29,14,6,11,20,13,26,4, 28,8,17**,31,**5,24,16,21,25, 23,30**,35,39,47,41,**9,3,19**,32,**1**} | 30 features with top weights: {**3,6,14,13,25,9,28,27,19,5,1, 16,20,8,18,2,10,29,11,26,12, 30,17,23,7,21,22,15,4,24**} |
| 2 | 26 features: {20:45} | 26 features with top weights: {**35,29,28,36,27,38,32,40,26, 31,25,37,39,34,30,33**,49,62, 80,129,89,127,190,150,186,120} | 26 features with top weights: {**35,29,28,36,27,38,32,40,26, 31,25,37,39,34,30,33**,49,62, 80,129,89,127,190,150,186,120} |
| 3 | 21 features: {35:55} | 21 features with top weights: {**38,42,51,52,44,47,43,41, 49,48,53,50,39,45,40,35, 55,36,54,37,46**} | 21 features with top weights: {**38,42,51,52,43,44,47,49,41, 48,50,53,39,45,40,35,55, 36,54,46,37**} |
| Identification Rate | | (24+16+21)/77=61/77 | (30+16+21)/77=67/77 |

The numbers in boldface denote the truly important features that were detected as being associated with the embedded subspaces.

Table 12 Comparison of the clustering performance of W-*k*-means, FSC, and ESSC on Toy-D.

| Index | | W-*k*-means | FSC | ESSC |
|---|---|---|---|---|
| *RI* | *mean* | 0.9708 | 0.9729 | **0.9870** |
| | *std* | 0.0117 | 0.0253 | **0.0210** |
| | * | 0.9842 | 1 | **1** |
| *NMI* | *mean* | 0.8727 | 0.9109 | **0.9574** |
| | *std* | 0.0327 | 0.0771 | **0.0686** |
| | * | 0.9080 | 1 | **1** |

* Denotes the values of RI and NMI achieved by each algorithm when the lowest value of the loss function is obtained within the 10 runs.

## 6 Model Complexity

Following the review of the three main categories of SSC algorithms, the computational complexity of the algorithms is discussed in this section. The complexities of the models of the representative SSC algorithms, as determined from the aspect of iterative rule based learning, are given in Table 13. It can be seen from the table that ISSC algorithms are usually computationally more complex than CSSC algorithms, but less complex than XSSC algorithms.

Table 13 Analysis of the models of the SSC algorithms

| Categories | Representative algorithms | Model analysis | | |
|---|---|---|---|---|
| | | Computational costs | Number of optimization variables | Initialization sensitivity |
| CSSC | WFCM [64] | $O(N+DN^2)$ | **3** | **Yes** |
| | W-*k*-means [39] | $O(TNC+TCD+TD)$ | **3** | **Yes** |
| ISSC | FSC [32] | $O(TNC+TC+TCD)$ | **3** | **Yes** |
| | EWKM [41] | $O(TNC+2TCD)$ | **3** | **Yes** |
| | EWMM [54] | $O(TNC+2TC+2TCD)$ | **5** | **Yes** |
| | COSA [28] | $O(NDL+TN^2D)$ * | **2** | **No** |
| XSSC | ESSC [20] | $O(TNC+2TCD)$ | **3** | **Yes** |
| | PSOVW [49] | $O(SDNCT)$ ** | **5** | **No** |
| | WFKCA [58] | $O(TNC+2TCD)$ | **3** | **Yes** |
| | TW-*k*-means [12] | $O(TCND)$ | **4** | **Yes** |

* $L$ is a predefined parameter of the nearest neighbors of a given sample. ** $S$ is the size of the particles for the PSO method.

## 7 Experimental Validation

To further appreciate the differences among the three categories of SSC algorithms, a representative algorithm is selected from each category to compare their clustering performance, i.e., the W-*k*-means algorithm of CSSC, the EWKM algorithm of ISSC, and the ESSC algorithm of XSSC. Their clustering performance is evaluated using high-dimensional gene expression datasets.

To measure the performance, the two classic clustering performance indices below are adopted:

(1) Rand index (RI):

$$RI = \frac{f_{00} + f_{11}}{N(N-1)/2},$$

where $f_{00}$ is the number of pairs of data points with different class labels and belonging to different clusters; $f_{11}$ is the number of pairs of data points with the same class labels and belonging to the same clusters; and $N$ is the size of the whole dataset.

(2) Normalized mutual information (NMI)

$$NMI = \frac{\sum_{i=1}^{C}\sum_{j=1}^{C} N_{i,j} \log \frac{N \cdot N_{i,j}}{N_i \cdot N_j}}{\sqrt{\sum_{i=1}^{C} N_i \log \frac{N_i}{N} \cdot \sum_{j=1}^{C} N_j \log \frac{N_j}{N}}},$$

where $N_{i,j}$ is the number of agreements between cluster $i$ and class $j$, $N_i$ is the number of data points in cluster $i$, $N_j$ is the number of data points in class $j$, and $N$ is the size of the whole dataset. Both RI and NMI take a value within the [0, 1] interval. The higher the value, the better the clustering performance. The parameter settings of the three subspace clustering algorithms are given in Table 14.

**Table 14 Parameter setting of the three subspace clustering algorithms.**

| Algorithm | Parameter Settings |
|---|---|
| W-*k*-means | $\beta = 2; 5; 10; 50; 100; 1000; 10^4; 10^5$ |
| EWKM | $\varepsilon_0 = 0, 10^{-10}, 10^{-5}, 10^{-4}, 10^{-3}, 10^{-2}$;<br>$\gamma = 1, 2, 5, 10, 50, 100, 1000$ |
| ESSC | $m = \frac{\min(N, D-1)}{\min(N, D-1) - 2}$;<br>$\gamma = 1, 2, 5, 10, 50, 100, 1000$;<br>$\eta = 0, 0.1, 0.2, 0.3, 0.5, 0.7, 0.9$ |

In all the experiments, the gene expression datasets are preprocessed by normalizing the feature in each dimension into the [0, 1] interval.

In the experiment, five real-world cancer gene expression datasets [5,7,55,56,66] are used to evaluate the performance of the three subspace clustering algorithms. The datasets are summarized in Table 15. Like many bioinformatics datasets, the cancer gene expression datasets used here contain a small number of samples with a large number of features, and thus suffer from the curse of dimensionality. The best clustering results, expressed in terms of the means and standard deviations of the RI and NMI values, and obtained by running each of the three algorithms 10 times, are recorded and shown in Tables 16 and 17.

It can be seen from the experimental results that the performance of the W-$k$-means algorithm (a CSSC algorithm) is inferior to that of EWKM, a classic ISSC algorithm. Meanwhile, ESSC (an XSSC algorithm) improves the clustering performance beyond the level achievable by the EWKM algorithm.

Table 15 Cancer gene expression datasets

| Dataset | Size of dataset | Number of dimensions | Number of clusters | Source |
|---|---|---|---|---|
| *CNS* | 34 | 7129 | 2 | [56] |
| *Prostate3* | 33 | 12626 | 2 | [66] |
| *Breast* | 84 | 9216 | 5 | [55] |
| *DLBCL* | 88 | 4026 | 6 | [5] |
| *Lung2* | 203 | 12600 | 5 | [7] |

Table 16 The best clustering results, in terms of RI, obtained using gene expression datasets.

| Dataset | | W-$k$-means | EWKM | ESSC |
|---|---|---|---|---|
| *CNS* | mean | 0.5086 | 0.5538 | 0.5805 |
| | std | 0.0172 | 0.0430 | 0.0578 |
| | * | 0.5134 | 0.5794 | 0.5989 |
| *Prostate3* | mean | 0.6924 | 0.7437 | 0.8573 |
| | std | 0.1772 | 0.1218 | 0.1422 |
| | * | 0.7348 | 0.8295 | 0.9394 |
| *Breast* | mean | 0.7203 | 0.7252 | 0.7318 |
| | std | 0.0286 | 0.0189 | 0.0166 |
| | * | 0.7510 | 0.7550 | 0.7728 |
| *DLBCL* | mean | 0.7654 | 0.7703 | 0.8585 |
| | std | 0.0411 | 0.0782 | 0.0212 |
| | * | 0.8161 | 0.8443 | 0.9206 |
| *Lung2* | mean | 0.5762 | 0.5871 | 0.7536 |
| | std | 0.0237 | 0.0114 | 0.0871 |
| | * | 0.5937 | 0.5970 | 0.6081 |

* Denotes the values achieved by each algorithm when the lowest value of the loss function is obtained within the 10 runs.

Table 17 The best clustering results, in terms of NMI, obtained using gene expression datasets.

| Dataset | | W-$k$-means | EWKM | ESSC |
|---|---|---|---|---|
| *CNS* | *mean* | 0.0924 | 0.1051 | 0.1312 |
| | *std* | 0.0688 | 0.0571 | 0.0870 |
| | * | 0.1216 | 0.1687 | 0.1721 |
| *Prostate3* | *mean* | 0.3978 | 0.4761 | 0.6889 |
| | *std* | 0.3110 | 0.0613 | 0.2150 |
| | * | 0.5754 | 0.6155 | 0.8130 |
| *Breast* | *mean* | 0.4312 | 0.4341 | 0.4917 |
| | *std* | 0.0498 | 0.0343 | 0.0994 |
| | * | 0.4596 | 0.4745 | 0.5526 |
| *DLBCL* | *mean* | 0.6005 | 0.6023 | 0.7526 |
| | *std* | 0.1055 | 0.0720 | 0.0683 |
| | * | 0.6825 | 0.7137 | 0.8527 |
| *Lung2* | *mean* | 0.2989 | 0.3433 | 0.5427 |
| | *std* | 0.0599 | 0.0294 | 0.0281 |
| | * | 0.3385 | 0.3628 | 0.3587 |

* Denotes the values achieved by each algorithm when the lowest value of the loss function is obtained within the 10 runs.

## 8  Conclusions

A comprehensive survey of SSC was presented in this paper. A wide variety of existing algorithms were systematically classified into three main categories: CSSC, ISSC, and XSSC. These three categories of SSC algorithms, along with the different subcategories, were reviewed and discussed in detail. This survey paper offers readers a thorough understanding of SSC algorithms and provides insights into future advancements in SSC. With an overall picture of SSC and a clear delineation of research developments on the subject, researchers can further conceptualize and conduct in-depth studies to address unresolved problems in SSC, and propose more adaptive and advanced SSC algorithms.

### Acknowledgments

This work was supported in part by the National Natural Science Foundation of China (61170122, 61272210), the Ministry of Education Program for New Century Excellent Talents (NCET-120882), the Fundamental Research Funds for Central Universities (JUSRP51321B), the Outstanding Youth Fund of Jiangsu Province (BK20140001), and the General Research Fund of the Hong Kong Research Grants Council (PolyU 5134/12E).

high-dimensional sparse data, IEEE Trans. Knowledge and Data Engineering 19(8) (2007) 1026-1041.

[42] L. Jing, M.K. Ng, J. Xu, et al., Subspace clustering of text documents with feature weighting k-means algorithm, In: Advances in Knowledge Discovery and Data Mining, Springer Berlin Heidelberg, 2005, pp. 802-812.

[43] A. Keller, F. Klawonn, Fuzzy clustering with weighting of data variables, International Journal of Uncertainty, Fuzziness and Knowledge-Based Systems 8(06) (2000) 735-746.

[44] H.P. Kriegel, P. Kröger, A. Zimek, Clustering high-dimensional data: A survey on subspace clustering, pattern-based clustering, and correlation clustering, ACM Transactions on Knowledge Discovery from Data 3(1) (2009) 1.

[45] H.P. Kriegel, P. Kröger, A. Zimek, Subspace clustering, Wiley Interdisciplinary Reviews: Data Mining and Knowledge Discovery 2(4) (2012) 351-364.

[46] T. Li, Y. Chen, An improved k-means algorithm for clustering using entropy weighting measures, in: Proceedings of the the 7th World Congress on Intelligent Control and Automation, IEEE, 2008, pp. 149-153.

[47] J. Li, X. Gao, L. Jiao, A new feature weighted fuzzy clustering algorithm, In: Rough Sets, Fuzzy Sets, Data Mining, and Granular Computing. Springer Berlin Heidelberg, 2005, pp. 412-420.

[48] T. Li, Y. Chen, A weight entropy k-means algorithm for clustering dataset with mixed numeric and categorical data, in: Fifth International Conference on Fuzzy Systems and Knowledge Discovery, IEEE, 2008, pp. 36-41.

[49] Y. Lu, S. Wang, S. Li, et al., Particle swarm optimizer for variable weighting in clustering high-dimensional data, Machine learning 82(1) (2011) 43-70.

[50] V. Makarenkov, P. Legendre, Optimal variable weighting for ultrametric and additive trees and K-means partitioning: Methods and software, Journal of Classification 18(2) (2001) 245-271.

[51] D.S. Modha, W.S. Spangler, Feature weighting in k-means clustering, Machine learning 52(3) (2003) 217-237.

[52] L. Parsons, E. Haque, H. Liu, Subspace clustering for high dimensional data: a review, ACM SIGKDD Explorations Newsletter 6(1) (2004) 90-105.

[53] H. Parvin, B. Minaei-Bidgoli, A clustering ensemble framework based on elite selection of weighted clusters, Advances in Data Analysis and Classification (2013) 1-28.

[54] L. Peng, J. Zhang, An entropy weighting mixture model for subspace clustering of high-dimensional data, Pattern Recognition Letters 32(8) (2011) 1154-1161.

[55] C.M. Perou, et al., Molecular Portraits of Human Breast Tumours, Nature 406 (2000) 747-752.

[56] S.L. Pomeroy, et al., Prediction of Central Nsystem Embryonal Tumour Outcome Based on Gene Expression, Nature 415(2002) 436-442.

[57] C.M. Procopiuc, M. Jones, P.K. Agarwal, et al., A Monte Carlo algorithm for fast projective clustering, in: Proceedings of the 2002 ACM SIGMOD International Conference on the Management of Data, ACM, 2002, pp. 418-427.

[58] H. Shen, J. Yang, S. Wang, et al., Attribute weighted mercer kernel based fuzzy clustering algorithm for general non-spherical datasets, Soft Computing 10(11) (2006) 1061-1073.

[59] K. Sim, V. Gopalkrishnan, A. Zimek, et al., A survey on enhanced subspace clustering, Data Mining and Knowledge Discovery 26(2) (2013) 332-397.

[60] M.E. Timmerman, E. Ceulemans, H.A. L. Kiers, et al., Factorial and reduced K-means reconsidered, Computational Statistics & Data Analysis 54(7) (2010) 1858-1871.

[61] C.Y. Tsai, C.C. Chiu, Developing a feature weight self-adjustment mechanism for a K-means clustering algorithm, Computational statistics & data analysis 52(10) (2008) 4658-4672.

[62] M. Vichi and H.A.L. Kiers, Factorial k-means analysis for two-way data, Computational Statistics & Data

Analysis 37(1) (2001) 49-64.

[63] J. Wang, Z.H. Deng, Y.Z. Jiang, P.J. Qian, and S.T. Wang, Multiple-kernel based soft subspace fuzzy clustering, in: Proceedings of 2014 IEEE International Conference on Fuzzy Systems (FUZZ-IEEE), IEEE, 2014, pp. 186-193.

[64] X. Wang, Y. Wang, L. Wang, Improving fuzzy c-means clustering based on feature-weight learning, Pattern Recognition Letters 25(10) (2004) 1123-1132.

[65] K.G. Woo, J.H. Lee, M.H. Kim, et al., FINDIT: a fast and intelligent subspace clustering algorithm using dimension voting, Information and Software Technology 46(4) (2004) 255-271.

[66] J.B. Welsh, Analysis of Gene Expression Identifies Candidate Markers and Pharmacological Targets in Prostate Cancer, Cancer Research 61 (2001) 5974-5978.

[67] J. Wang, F.L. Chung, S.T. Wang, and Z.H. Deng, Double indices-induced FCM clustering and its integration with fuzzy subspace clustering, Pattern Analysis and Applications 17(3) (2014) 549-566.

[68] Y. Xiao, J. Yu, Partitive clustering (K-means family), Wiley Interdisciplinary Reviews: Data Mining and Knowledge Discovery 2(3) (2012) 209-225.

[69] R. Xu, D. Wunsch, Survey of clustering algorithms, IEEE Trans. Neural Networks 16(3) (2005) 645-678.

[70] J. Yang, W. Wang, H. Wang, et al., δ-clusters: Capturing subspace correlation in a large data set, in: Proceedings of the 18th International Conference on Data Engineering, IEEE, 2002, pp. 517-528.

[71] K.Y. Yip, D.W. Cheung, M.K. Ng, Harp: A practical projected clustering algorithm, IEEE Trans. Knowledge and Data Engineering 16(11) (2004) 1387-1397.

[72] L. Zhu, L. Cao, J. Yang, Multiobjective evolutionary algorithm-based soft subspace clustering, in: 2012 IEEE Congress on Evolutionary Computation, IEEE, 2012, pp. 1-8.